\makeatletter\def\input@path{{./}{Arxiv/}}\makeatother
\documentclass{article}
\usepackage{arxiv}
\usepackage[normalem]{ulem}
\usepackage{cite}
\usepackage{amsmath,amssymb,amsfonts,amsthm}
\usepackage{graphicx,color}
\usepackage{xcolor}
\graphicspath{{./}{Arxiv/}}
\usepackage{array}
\usepackage{url}
\usepackage{algorithm}
\usepackage{algpseudocode}
\usepackage{hyperref}
\hypersetup{colorlinks=true,linkcolor=black,citecolor=black,
  urlcolor=black,
  pdftitle={Capability Composition in Heterogeneous Teams via Barrier-Guided
    Diffusion Sampling},
  pdfauthor={Diana Laura Morales, Ruoyu Lin, Magnus Egerstedt, Fabio
    Pasqualetti}}

\newtheorem{theorem}{Theorem}[section]
\newtheorem{lemma}[theorem]{Lemma}
\newtheorem{proposition}[theorem]{Proposition}
\newtheorem{corollary}[theorem]{Corollary}
\newtheorem{assumption}[theorem]{Assumption}

\newtheorem{example}{Example}

\newcommand{\real}{\mathbb{R}}
\newcommand{\N}{\mathcal{N}}

\newcommand{\supp}{\mathrm{supp}}
\newcommand{\smin}{\mathrm{smin}}
\newcommand{\smax}{\mathrm{smax}}
\newcommand{\ccoup}[1]{c_{\mathrm{coup},#1}}
\newcommand{\ctask}{c_{\mathrm{task}}}
\newcommand{\norm}[1]{\left\lVert #1 \right\rVert}

\newcommand\oprocendsymbol{\hbox{$\square$}}
\newcommand\oprocend{\relax\ifmmode\else\unskip\hfill\fi\oprocendsymbol}

\title{Capability Composition in Heterogeneous Teams via Barrier-Guided Diffusion Sampling}

\renewcommand{\shorttitle}{Capability Composition via Barrier-Guided Diffusion}
\renewcommand{\undertitle}{}
\renewcommand{\headeright}{}

\author{%
  Diana Laura Morales \\
  Electrical Engineering and Computer Science\\
  University of California, Irvine\\
  \texttt{dlmoral2@uci.edu}
  \And
  Ruoyu Lin \\
  Electrical Engineering and Computer Science\\
  University of California, Irvine\\
  \texttt{rlin10@uci.edu}
  \AND
  Magnus Egerstedt \\
  University of North Carolina at Chapel Hill\\
  \texttt{magnus@unc.edu}
  \And
  Fabio Pasqualetti \\
  Electrical Engineering and Computer Science\\
  University of California, Irvine\\
  \texttt{fabiopas@uci.edu}
}

\begin{document}
\maketitle

\begin{abstract}
The agents in a heterogeneous team can cooperate to accomplish tasks that exceed their individual capabilities. The team trajectories for a cooperative task can be generated by a generative model, provided that the model has been trained on demonstrations of the agents cooperating. Yet, demonstrations of cooperation are costly and may not exist for tasks that the team has never performed. The conditions under which the agents can combine their capabilities, however, are physical properties of the team and are known before any demonstration is collected. We encode the conditions as constraints on the team trajectories, and we enforce the constraints while sampling from generative models trained on demonstrations of each agent alone. We prove that the proposed method can generate every team trajectory that satisfies the constraints, although the demonstrations contain no cooperation. Hence the constraints determine which team trajectories can be generated, and the demonstrations determine how likely each team trajectory is. Numerical results on two cooperative tasks show the effectiveness of the method, which generates cooperative team trajectories that satisfy the constraints and requires no retraining when agents join or leave the team. Finally, we implement the method on a multi-robot testbed.
\end{abstract}


\section{Introduction}
\label{sec:intro}

Cooperation allows the agents in a team to accomplish tasks that exceed their individual capabilities. Cooperation in multi-agent teams has long been studied, from the allocation and scheduling of tasks~\cite{BG-MM:04, GAK-AS-MBD:13} to the control of several robots acting on a common object~\cite{ET-MHMA-OA:18, HF-SB:22}. Existing methods assume that the capabilities of the team are specified in advance, together with the conditions under which each capability can be used. The capabilities that a task requires, however, depend on the environment where the agents are deployed, and the environment is often unknown when the agents are trained. The task may then require a capability that no agent has alone, and that the agents obtain only by combining their individual capabilities. Hence the required cooperation can be neither specified nor demonstrated in advance, and generating cooperative behaviors for the task remains a challenge.

Generative models, and diffusion models in particular, are effective methods to generate the trajectories of an agent~\cite{MJ-YD-JBT-SL:22, CC-ZX-SF-EC-YD-BB-RT-SS:23} or of a team~\cite{ZZ-ML-LM-BK-MX-YY-SE-WZ:24, DL-DL-AZ:26}, provided that the demonstrations contain the desired behavior. 
To overcome this limitation, we observe that the conditions under which the agents can combine their capabilities follow from the physics of the team. Hence the conditions can be written as constraints on the team trajectories without any demonstration. We then sample the team trajectories from generative models trained on demonstrations of each agent alone, and we impose the constraints during sampling using a guidance method~\cite{PD-AN:21, MJ-YD-JBT-SL:22}. This results in a method to sample cooperative team trajectories from models pre-trained on non-cooperative demonstrations.

\noindent
\textbf{Related work.} Task allocation and scheduling decide which agent performs which task and when~\cite{BG-MM:04, GAK-AS-MBD:13}, and a trajectory
planner then realizes the decision. We sample the trajectories directly from diffusion models and read the pattern of cooperation off the result, with
nothing computed in advance. One line of multi-agent diffusion planners trains on demonstrations of cooperation where they exist~\cite{ZZ-ML-LM-BK-MX-YY-SE-WZ:24, DL-DL-AZ:26}, and reproduces only the cooperation those demonstrations contain. Another trains on single-robot data and reconciles the resulting plans by search~\cite{YS-IM-SV-JL-ML:25} or by projection~\cite{JL-JKC-SK-FF:25}. These approaches combine constraints by intersection, as for collision avoidance. We combine them by union, because cooperation admits configurations that no agent reaches alone. Cooperation is available only while the agents are physically arranged for it, and has to be encoded as a constraint that holds at every timestep of a trajectory. Control barrier functions~\cite{ADA-XX-JWG-PT:17} (barriers, for short) encode such constraints. A barrier keeps a trajectory inside the set it defines. Boolean combinations of barriers give intersections and unions of the admissible sets~\cite{PG-JC-ME:20,RL-ME:25}. On the generative side, barriers have been used to keep 
diffusion samples inside a safe set~\cite{WX-THW-CG-DR:25} and to guide sampling with a constricting condition that tightens as noise falls~\cite{DG-AA-FP:26}. Barriers also make flow-matching samplers safe~\cite{XD-ZY-DY-FL-HS-SH-SH:25,JY-SJ-SH:26}. In multi-agent systems, guidance can be distributed, so that each agent computes its own Lyapunov-based guidance input~\cite{RL-ME-FP:26}.

\noindent
\textbf{Contributions.} Our contributions are as follows. First, we formulate cooperative trajectory generation without demonstrations of cooperation as barrier-guided diffusion sampling, with constraints that encode the individual capabilities of the agents and the cooperative ones. Second, we propose a scheme in which the scores trained independently for each agent are summed into a product prior, and we prove that the sum is the exact
score of that product at every noise level, since the densities of different agents depend
on disjoint sets of variables. Third, we prove that the sampler can generate
every team trajectory satisfying the constraints with nonzero probability
and, when its guidance succeeds at every step, no other. Thus, every pattern of cooperation the constraints permit can be
generated, and none is specified or pre-computed. Fourth, we demonstrate the method on a river crossing
task, in simulation and on the Robotarium. The generated trajectories cross the river cooperatively, while sampling
the prior alone yields agents that ignore each other and fail the task. We also show that the sampler requires no retraining
to generate trajectories when agents join or leave the team. Finally,
we consider a task in which cooperation also has to be timed. The sampler generates such trajectories without demonstrations, but less often than in the river crossing. Most failures of the sampler are consistent with local maxima of the barrier.

\section{Problem formulation}\label{sec:problem}
Consider a team of $N$ agents with states $x_i \in \real^{n_i}$ and inputs
$u_i \in \real^{m_i}$,
evolving in discrete time as
\begin{equation}
  x_i(t+1) = f_i\big(x_i(t), u_i(t)\big), \quad t = 0, \dots, T-1.
  \label{eq:dyn}
\end{equation}
Let $x = (x_1, \dots, x_N) \in \real^{n}$ be the team state and
$u = (u_1, \dots, u_N) \in \real^{m}$ the team input, with $n = \sum_i n_i$ and
$m = \sum_i m_i$. A team trajectory is the matrix
$X = \big(x(0), \dots, x(T)\big) \in \real^{n \times (T+1)}$, whose row block
$X_i \in \real^{n_i \times (T+1)}$ corresponds to the trajectory of agent $i$.

Without cooperation, agent $i$ is constrained to the states and controls
satisfying $c_i(x_i,u_i) \ge 0$, where
$c_i : \real^{n_i} \times \real^{m_i} \to \real$ is a problem-dependent
constraint function. Cooperation enlarges this set. Let
$\ccoup{i} : \real^{n} \times \real^{m} \to \real$ describe the allowable
configurations while cooperating with the rest of the team, so that the
constraint set of agent $i$ is
\begin{equation}\label{eq:union}
  \mathcal{A}_i = \{(x,u) : c_i(x_i,u_i) \ge 0\} \cup
  \{(x,u) : \ccoup{i}(x,u) \ge 0\}.
\end{equation}
A team trajectory is admissible when the pair $(x(t), u(t))$ lies in
$\mathcal{A}_i$ for every agent $i$ and at all times. Cooperation enlarges the
set of admissible states and controls, while the dynamics \eqref{eq:dyn} remain
the same whether or not $\ccoup{i} \ge 0$. Combining constraint sets is standard in
the safe control literature~\cite{PG-JC-ME:20}. Multi-robot
planning combines them by intersection, since collision avoidance asks that a
plan satisfy every agent's constraints, and the feasible set may
shrink~\cite{YS-IM-SV-JL-ML:25}. In \eqref{eq:union}, instead, the second set
depends on the other agents, so $\mathcal{A}_i$ admits configurations that agent
$i$ cannot reach alone.

A task is specified by a problem-dependent function
$\ctask : \real^{n \times (T+1)} \to \real$, and a team trajectory successfully
completes the task when $\ctask(X) \ge 0$. We are interested in tasks that no
agent can complete alone and the team can complete together. This motivates the
following assumption.

\begin{assumption}{\emph{\textbf{(Cooperation is necessary and
        sufficient to complete the task)}}}\label{as:coop}
  Every admissible team trajectory $X$ that completes the task also satisfies
  $c_i(x_i(t),u_i(t)) < 0 \le \ccoup{i}(x(t),u(t))$ for some agent $i$
  and some time $t$. Moreover, at least one such trajectory exists.~\oprocend
\end{assumption}

Assumption \ref{as:coop} requires that, along every trajectory that completes the task, some agent at some timestep violates its own constraint
$c_i$ and is admissible only through the coupling constraint $\ccoup{i}$. Hence no trajectory in which every agent always satisfies its own constraint completes the task, that is, cooperation is necessary. Cooperation is also sufficient, because at least one admissible trajectory completes the task.

\begin{figure}[t]
\centering
\includegraphics[width=0.6\textwidth]{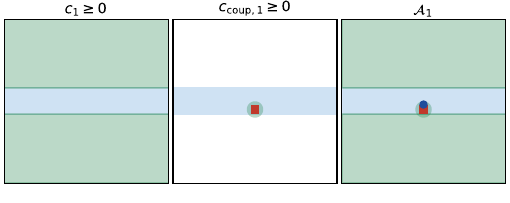}
\caption{Positions admitted by the vehicle's constraint set
  $\mathcal{A}_1$ in Example~\ref{ex:ferry}, for one position of the vessel
  (square): the land, by $c_1$ (left), within $r$ of 
  the vessel, by $\ccoup{1}$ (middle), and their union (right).
  The disc is narrower than the river, so it reaches land only when the
  vessel is near it. The vehicle (circle) is aboard the vessel when
  $c_1 < 0$ and $\ccoup{1} \ge 0$.}
\label{fig:concept}
\end{figure}

\begin{example}{\emph{\textbf{(Crossing a river)}}}\label{ex:ferry}
  Consider a team of two agents, a ground vehicle and a vessel, operating in a
  unit square environment divided across the middle by a river of width $w$. The
  task is for the vehicle to reach a location across the river. The vehicle
  cannot enter the water and the vessel cannot leave it, so the vehicle cannot
  satisfy the task alone.

  Let $x_i = (z_i, q_i)$ be the position of agent $i$, with $z_i$ and $q_i$ as
  its horizontal and vertical coordinates, respectively. Agent $i=1$ is the
  vehicle and agent $i=2$ is the vessel, with dynamics $f_i(x_i,u_i)=x_i+u_i$.
  The constraint functions are
  \begin{align*}
    c_1(x_1,u_1) &= \min\big( |q_1 - \tfrac{1}{2}| - \tfrac{w}{2},\;
                             v^{\max} - \norm{u_1} \big),\\
    c_2(x_2,u_2) &= \min\big( \tfrac{w}{2} - |q_2 - \tfrac{1}{2}|,\;
                             v^{\max} - \norm{u_2} \big),\\
    \ccoup{1}(x,u) &= \min\big( r - \norm{x_1 - x_2},\;
                               v^{\max} - \norm{u_1 - u_2} \big),
  \end{align*}
  and $\ccoup{2} < 0$ everywhere, since cooperation never lets the vessel leave
  the water. Hence $c_1 \ge 0$ when the vehicle is on land and within its own
  speed limit, $c_2 \ge 0$ when the vessel is on water and within its own, and
  $\ccoup{1} \ge 0$ when the two are within a tolerance radius $r$ and the
  vehicle moves within $v^{\max}$ of the vessel. Figure \ref{fig:concept} shows
  these sets (positions). The vehicle may therefore be in the water only while
  riding the vessel. While the vehicle is aboard, its speed limit applies relative
  to the vessel, so its velocity is the velocity of the vessel plus an input of
  norm at most $v^{\max}$. Since
  $\ccoup{2} < 0$ leaves
  $\mathcal{A}_2 = \{(x,u) : c_2(x_2,u_2) \ge 0\}$, the vessel is bound by
  $\norm{u_2} \le v^{\max}$ at all times, and
  $\norm{u_1} \le 2 v^{\max}$ whenever the vehicle is within $r$ of the
  vessel. The vehicle starts at a given position $x_1^0$ on one side of the
  river, the vessel at a given position $x_2^0$ on the water, and the vehicle
  must reach a goal $g$ across the river:
  \begin{equation*}
    \ctask(X) = \min\big( \delta - \norm{x_1(0) - x_1^0},\,
                          \delta - \norm{x_2(0) - x_2^0},\,
                          \delta - \norm{x_1(T) - g} \big),
  \end{equation*}
  where $\delta >0$ is the tolerance on each of the three positions.~\oprocend
\end{example}

We study the following \emph{capability composition problem}: given a task
$\ctask$ and a team of $N$ agents with constraint sets $\mathcal{A}_i$, sample
team trajectories $X$ with $\ctask(X) \ge 0$ and
$(x(t),u(t)) \in \mathcal{A}_i$ for every $i$ and every $t$. We generate
trajectories with a denoising diffusion model, a dynamical system that
transports a Gaussian into the data distribution~\cite{JH-AJ-PA:20, YS-JSD-DPK-AK-SE-BP:21}. A forward process adds noise to $X$ over diffusion
time $\tau \in [0,1]$ as
\begin{equation}
  \mathrm{d}X = -\tfrac{1}{2}\,g^2(\tau)\, X \,\mathrm{d}\tau
              + g(\tau)\,\mathrm{d}w,
  \label{eq:forward}
\end{equation}
where $g > 0$ is the noise schedule and $w$ is a Wiener process. We write $p_\tau$ for
the density of $X$ at time $\tau$, so that $p_0$ is the density of the
demonstrations and $p_1$ close to the standard Gaussian $\N(0,I)$. The coefficients of
\eqref{eq:forward} are scalar, so noise enters each coordinate
independently. Sampling reverses \eqref{eq:forward}, running from $\tau = 1$
back to $\tau = 0$, according to
\begin{equation}
  \mathrm{d}X = f_\theta(X,\tau)\,\mathrm{d}\tau + g(\tau)\,\mathrm{d}w,
  \label{eq:sde}
\end{equation}
with $\mathrm{d}w$ now a reverse-time Wiener process and the drift $f_\theta$
built from a score network trained to approximate the score
$\nabla_X \log p_\tau$.
We assume that no demonstration of cooperation is available for training.
Let $\mathcal{D}_i$ be a
set of demonstrations $X_i \in \real^{n_i \times (T+1)}$ of agent $i$ alone,
let $p^i$ be their density, and let $s_{\theta_i}$ be a score network trained on
$\mathcal{D}_i$. The team prior is the product $p^\otimes = \prod_i p^i$ of the
per-agent densities. We assume that each $\mathcal{D}_i$ is collected in an
unconstrained environment, in which agent $i$ is free to occupy any
state. The demonstrations are therefore
reusable across tasks, while $c_i$ is task-specific and
enforced at sampling time. In Example \ref{ex:ferry}, $\mathcal{D}_1$ and
$\mathcal{D}_2$ contain trajectories of the vehicle and the vessel.

\section{Main results}\label{sec:main}
Generating a team trajectory through \eqref{eq:sde} requires the team score at
every diffusion time, and Section \ref{sec:problem} leaves us with one score per
agent. Let $p^\otimes_\tau$ denote the density at time $\tau$ of
the forward process \eqref{eq:forward} started from $p^\otimes$. The separately
trained per-agent scores yield the team score exactly, as we show next.

\begin{proposition}{\emph{\textbf{(Agent scores form the team
        score)}}}\label{prop:product}
  Let $p^i_\tau$ be the density of $X_i$ at time $\tau$. Then the rows of the
  team score for agent $i$ are its own score: for every $\tau$ and $i$,
  \begin{equation}\label{eq:product}
    \nabla_{X_i} \log p^\otimes_\tau(X) = \nabla_{X_i} \log p^i_\tau(X_i).
  \end{equation}
\end{proposition}

\begin{proof}
  Let $k_\tau(X \mid X_0)$ be the density of $X$ at time $\tau$ when the forward
  process starts from the trajectory $X_0$. It factorizes across coordinates, so
  $p^\otimes_\tau(X) = \prod_i \int p^i(Y_i) \, k_\tau(X_i \mid Y_i) \,
  \mathrm{d}Y_i$, and the logarithm turns the product into a sum whose $i$-th
  term depends on $X_i$ alone.
\end{proof}
The identity \eqref{eq:product} holds whenever the densities in the product depend
on disjoint sets of variables. The forward process adds noise to each coordinate
independently, so it smooths each variable separately. When the densities are
over the \emph{same} variable, as in a product of experts, smoothing the product
is not the same as multiplying the smoothed densities, and the identity holds
only approximately~\cite{YD-CD-RS-JBT-SD-RF-JSD-AD-WG:23}.

Sampling \eqref{eq:sde} with this score returns team trajectories drawn from
$p^\otimes$, and there is no reason for them to be admissible or to complete the
task: the per-agent densities were trained on demonstrations and carry no
information about $\mathcal{A}_i$ or $\ctask$. The constraints must therefore be imposed while the trajectory is
generated.
Following~\cite{DG-AA-FP:26}, we add a control $u$ to \eqref{eq:sde},
\begin{equation}\label{eq:gsde}
  \mathrm{d}X = \big[ f_\theta(X,\tau) + u \big]\,\mathrm{d}\tau
              + g(\tau)\,\mathrm{d}w,
\end{equation}
and choose $u$ at each step so that a scalar barrier stays non-negative along
the sampled trajectory. The barrier encodes all the constraints specified in
Section \ref{sec:problem}. The sampler generates state trajectories only, while the
constraint functions depend also on the inputs. We therefore assume that, for
every agent $i$ and state $x_i$, the map $u_i \mapsto f_i(x_i,u_i)$ in
\eqref{eq:dyn} is injective, so that $u_i(t)$ is the unique input taking
$x_i(t)$ to $x_i(t+1)$. In Example~\ref{ex:ferry},
$u_i(t) = x_i(t+1) - x_i(t)$. Hence the constraints along a
trajectory are functions of $X$ only. Admissibility is an intersection over agents and times, and
\eqref{eq:union} is a union:
\begin{equation}
  \underbrace{\min_{i,t}\, \max\big\{ c_i(x_i(t),u_i(t)),\;
    \ccoup{i}(x(t),u(t)) \big\}}_{h_{\mathrm{comp}}(X)},
  \label{eq:hcomp}
\end{equation}
so that $h_{\mathrm{comp}} \ge 0$ when the trajectory is admissible. The
actuation limits do not need a term of their own, since \eqref{eq:union} already
carries each agent's actuation limit in $c_i$ and the coupled regime's in
$\ccoup{i}$. The barrier then becomes
\begin{equation*}
  h(X) = \min\{h_{\mathrm{comp}}(X), \ctask(X)\},
\end{equation*}
and the set $\mathcal{C} = \{X: h(X) \ge 0\}$ contains exactly the admissible team trajectories that complete the task.

The sampler of~\cite{DG-AA-FP:26} simulates \eqref{eq:gsde} on a grid
$\tau_K > \dots > \tau_0 = 0$ with steps $\Delta\tau_k = \tau_k - \tau_{k-1}$,
\begin{equation}\label{eq:step}
  X_{k-1} = X_k - \big[ f_\theta(X_k,\tau_k) + u_k \big]\Delta\tau_k
            + g(\tau_k)\sqrt{\Delta\tau_k}\,\zeta_k ,
\end{equation}
from $X_K \sim \N(0,I)$ to the returned trajectory $X^\star = X_0$, where the
$\zeta_k$ are independent Gaussian increments. The control uses the
relaxed barrier $\tilde h(X,\tau) = h(X) + \epsilon(X_K,\tau)$, whose margin is
continuous, nondecreasing in $\tau$, vanishes at $\tau_0$, and equals $\max\{0,-h(X_K)\}$
at $\tau_K$, so that $\tilde h(X_K,\tau_K) \ge 0$. At each step $\zeta_k$ is
drawn first, then $u_k$ is chosen so that
\begin{equation}\label{eq:cbf}
  \tilde h(X_{k-1},\tau_{k-1}) \ge (1 - \alpha\Delta\tau_k)\,
  \tilde h(X_k,\tau_k),
\end{equation}
with $\alpha > 0$ and $\alpha\Delta\tau_k \le 1$. If \eqref{eq:cbf} holds at
every step, then $\tilde h$ stays non-negative and $h(X^\star) \ge 0$ for every 
realization of the noise~\cite{DG-AA-FP:26}. 
In that case $X^\star$ is admissible and completes the task, and, by Assumption
\ref{as:coop}, at some timestep some agent violates its own
constraint $c_i$ and is admissible only through its coupling constraint
$\ccoup{i}$. 
Let $X_{k-1}(u)$ denote the state that \eqref{eq:step} gives with $u_k = u$ and the noise $\zeta_k$ already drawn, so that \eqref{eq:cbf} depends on $u_k$ through $\tilde h(X_{k-1}(u_k),\tau_{k-1})$.
Linearizing $\tilde h(\cdot,\tau_{k-1})$ at a point $\bar X$ gives the
condition
\begin{equation}\label{eq:lin}
  \tilde h(\bar X,\tau_{k-1})
  + \big\langle \nabla\tilde h(\bar X,\tau_{k-1}),\, X_{k-1}(u) - \bar X \big\rangle
  \ge (1 - \alpha\Delta\tau_k)\,\tilde h(X_k,\tau_k),
\end{equation}
which is linear in $u$, so the control of minimum norm that satisfies \eqref{eq:lin} is in closed form. The sampler of~\cite{DG-AA-FP:26} takes the minimum-norm control with
$\bar X = X_k$. As noted in~\cite{DG-AA-FP:26}, the linearization error is quadratic in the
displacement $X_{k-1} - X_k$, which is dominated by the noise term
$g(\tau_k)\sqrt{\Delta\tau_k}\,\zeta_k$. The error is therefore of order
$g(\tau_k)^2\Delta\tau_k$, the same order as the decrease that \eqref{eq:cbf}
allows. Hence a control that satisfies \eqref{eq:lin} at $\bar X = X_k$ can violate \eqref{eq:cbf}, and the guarantee is then lost.

To overcome this limitation, we use the sampler of~\cite{DG-AA-FP:26} with three
modifications. First, the relaxed barrier is evaluated on an estimate
of the trajectory being formed rather than on the iterate itself, which for
most of the reverse process is dominated by noise and carries little
information about the trajectory the sampler returns. The
estimate is the best affine estimate of $X_0$ from $X_\tau$, given in
Lemma~\ref{lem:Mtau} below. Second, the control is $u_k = 0$
whenever the unguided update satisfies \eqref{eq:cbf}, instead of its
linearization as in~\cite[Remark~2]{DG-AA-FP:26}. 
Third, when the unguided update violates \eqref{eq:cbf}, we search for
$u_k$ locally. We first set $u_k$ to the control of minimum norm that satisfies
\eqref{eq:lin} with $\bar X = X_k$, as in~\cite{DG-AA-FP:26}. While
\eqref{eq:cbf} fails, we set $\bar X = X_{k-1}(u_k)$, the state reached with
the current control, and recompute $u_k$ from \eqref{eq:lin}. If no control
satisfying \eqref{eq:cbf} is found within $j_{\max}$ iterations, as happens
near a local maximum of $\tilde h$, where the gradient vanishes, we take the
step with the last control. The guarantee is then lost, but the returned
trajectory may still lie in $\mathcal{C}$, because later steps can increase
$\tilde h$ again. For this reason, Algorithm~\ref{alg:sampler} returns
$h(X^\star)$ together with $X^\star$.

\begin{lemma}{\emph{\textbf{(Affine estimate of the returned
      trajectory)}}}\label{lem:Mtau}
  Let $X_0 \in \real^{n \times (T+1)}$ be a random variable with mean $\mu$
  and covariance $\Sigma > 0$. Let $X_\tau$ evolve from $X_0$ according to \eqref{eq:forward},
  and let $a_\tau = \exp\!\big(-\tfrac{1}{2}\int_0^\tau g^2(s)\,\mathrm{d}s\big)$
  and $\sigma_\tau^2 = 1 - a_\tau^2$. Then
  \begin{equation*}
    (A_\tau, b_\tau) = \arg\min_{A,\, b}\, \mathbb{E}\norm{X_0 - A X_\tau - b}^2
  \end{equation*}
  is given by $A_\tau = a_\tau \Sigma\,\big(a_\tau^2 \Sigma + \sigma_\tau^2 I\big)^{-1}$
  and $b_\tau = (I - a_\tau A_\tau)\mu$, so that $M_\tau X = A_\tau X + b_\tau$
  satisfies
  \begin{equation}\label{eq:Mtau}
    M_\tau X = \mu + A_\tau (X - a_\tau \mu).
  \end{equation}
  Moreover, $M_\tau$ is invertible for every $\tau \in [0,1]$, $M_0 = I$, and $M_\tau X = \mathbb{E}[X_0 \mid X_\tau = X]$ when $X_0$ is
  Gaussian.
\end{lemma}
\begin{proof}
  Equation \eqref{eq:forward} is linear with scalar coefficients, so
  $X_\tau = a_\tau X_0 + \sigma_\tau \xi$, where $\xi \sim \N(0,I)$ is
  independent of $X_0$ and
  $\sigma_\tau^2 = a_\tau^2\int_0^\tau g^2(s)/a_s^2\,\mathrm{d}s = 1 - a_\tau^2$.
  Hence $X_\tau$ has mean $a_\tau\mu$ and covariance
  $a_\tau^2\Sigma + \sigma_\tau^2 I$, and its cross-covariance with $X_0$ is
  $a_\tau\Sigma$. The minimizer is
  $A_\tau = \mathrm{Cov}(X_0,X_\tau)\,\mathrm{Cov}(X_\tau)^{-1}$ and
  $b_\tau = \mathbb{E}[X_0] - A_\tau\,\mathbb{E}[X_\tau]$, which gives the
  expressions above. The eigenvalues $\lambda_j$ of $\Sigma$ are positive, and
  $A_\tau$ has the same eigenvectors and the eigenvalues
  $a_\tau\lambda_j/(a_\tau^2\lambda_j + \sigma_\tau^2) > 0$, so $M_\tau$ is
  invertible. Since $a_0 = 1$ and $\sigma_0 = 0$, $A_0 = I$ and $b_0 = 0$.
  When $X_0$ is Gaussian, $(X_0, X_\tau)$ is jointly Gaussian, and its
  conditional mean is affine and equal to \eqref{eq:Mtau}.
\end{proof}

Lemma~\ref{lem:Mtau} uses only the mean and the covariance of $X_0$, so
$M_\tau$ is computed in closed form and does not need to be learned. During
sampling, $X_\tau$ is the current iterate, and $M_\tau X_\tau$ is the best
affine estimate of the trajectory being formed. We therefore let $\mu$ and
$\Sigma$ be the mean and covariance of the product prior $p^\otimes$, estimated
from the training sets $\mathcal{D}_i$, and evaluate the relaxed barrier on
this estimate, that is,
$\tilde h(X,\tau) = h(M_\tau X) + \epsilon(X_K,\tau)$, with margin
$\epsilon(X_K,\tau_K) = \max\{0,-h(M_{\tau_K}X_K)\}$. Theorem~4.2 of~\cite{DG-AA-FP:26} is stated for the barrier
$h(X) + \epsilon(X_K,\tau)$, and it extends to the modified $\tilde h$ above with the same
proof. The margin gives $\tilde h(X_K,\tau_K) \ge 0$, and if \eqref{eq:cbf}
holds at every step, then
$\tilde h(X_{k-1},\tau_{k-1}) \ge (1-\alpha\Delta\tau_k)\,\tilde h(X_k,\tau_k)
\ge 0$ for every $k$, by induction. As $\epsilon$ vanishes at $\tau_0$ and
$M_0 = I$, $h(X^\star) = \tilde h(X_0,\tau_0) \ge 0$, so $X^\star \in \mathcal{C}$. In our implementation, the right-hand side of \eqref{eq:cbf} is replaced
by its minimum with $0.15 + \epsilon(X_K,\tau_{k-1})$, so that the guidance
stops acting once a trajectory is well inside $\mathcal{C}$. The right-hand
side remains non-negative, and the argument above still applies.

\begin{algorithm}[t]
  \caption{Barrier-guided sampling, modified from~\cite{DG-AA-FP:26}}
  \label{alg:sampler}
  \small
  \begin{algorithmic}
    \Require per-agent scores $s_{\theta_i}$, relaxed barrier $\tilde h$, grid
      $\{\tau_k\}$, gain $\alpha$, iteration limit $j_{\max}$
    \State Draw $X_K \sim \N(0,I)$ and set the margin $\epsilon(X_K,\cdot)$
    \For{$k = K, \dots, 1$}
      \State Draw $\zeta_k \sim \N(0,I)$ and set $u_k \gets 0$
      \If{$X_{k-1}(0)$ violates \eqref{eq:cbf}}
        \State $u_k \gets \arg\min\{\norm{u} : \text{\eqref{eq:lin}},\ \bar X = X_k\}$~\cite{DG-AA-FP:26}
        \For{$j = 1, \dots, j_{\max}$, while \eqref{eq:cbf} fails at $X_{k-1}(u_k)$}
          \State $u_k \gets \arg\min\{\norm{u} : \text{\eqref{eq:lin}},\ \bar X = X_{k-1}(u_k)\}$
        \EndFor
      \EndIf
      \State $X_{k-1} \gets X_{k-1}(u_k)$
    \EndFor
    \State \Return $X^\star = X_0$ and $h(X^\star)$
  \end{algorithmic}
\end{algorithm}

We next characterize the type of trajectories that the diffusion sampler can
generate. In particular, we prove that any admissible team trajectory solving
the task can be generated with arbitrary accuracy and nonzero probability, even
when the training set does not contain such trajectories. To this end, we make
the following technical assumption.

\begin{assumption}{\emph{\textbf{(Monotone reachability)}}}\label{as:mono}
  For every $X \in \mathcal{C}$ and $s \le 0$, there exists $X'$ with
  $h(X') = s$ and a continuous path from $X'$ to $X$ along which $h$
  is~nondecreasing.~\oprocend
\end{assumption}

Assumption \ref{as:mono} asks that every trajectory in $\mathcal{C}$ be reachable from
below without lowering $h$. It fails when a trajectory in $\mathcal{C}$ lies in a
region enclosed by higher values of $h$, so that $h$ first increases and then
decreases along every path that reaches it from a lower level set. It involves only $h$, and neither the prior nor the data. For
$X \in \real^{n \times (T+1)}$ and $\rho > 0$, $B(X,\rho)$ denotes the open
ball of radius $\rho$ centered at $X$ in the Frobenius norm, and $\supp(\mu)$
denotes the support of a distribution $\mu$, that is, the smallest closed
set of probability~one.

\begin{theorem}{\emph{\textbf{(Every trajectory in $\mathcal{C}$ can be
        generated)}}}\label{thm:complete}
  Let Assumption~\ref{as:mono} hold, let $f_\theta$ and $h$ be continuous,
  with $\mathcal{C}$ the closure of $\{X : h(X) > 0\}$, and let $\mu$ be the
  distribution of the trajectory $X^\star$ returned by
  Algorithm~\ref{alg:sampler}. Then
  $\mu\big(B(\bar X,\rho) \cap \mathcal{C}\big) > 0$ for every
  $\bar X \in \mathcal{C}$ and every $\rho > 0$. If, in addition,
  \eqref{eq:cbf} holds at every step, then $\supp(\mu) = \mathcal{C}$.
\end{theorem}
\begin{proof}
Fix $\bar X$ with $h(\bar X) > 0$ and $\rho > 0$, and let $\rho' \in (0,\rho]$ be such that
$h > 0$ on $B(\bar X,\rho')$, which exists because $h$ is
continuous. Assumption \ref{as:mono} with
any $s < 0$ gives a trajectory $\hat Y_K$ with $h(\hat Y_K) = s$ and a
continuous path from $\hat Y_K$ to $\bar X$ along which $h$ is
nondecreasing. Let $\hat X_K = M_{\tau_K}^{-1}\hat Y_K$. At
$\hat X_K$ the margin is $\epsilon(\hat X_K,\tau_K) = -s$, so
$\tilde h(\hat X_K,\tau_K) = 0$, and as $\epsilon$ vanishes at $\tau_0$ and
$h(\bar X) > 0$, $\textstyle\sum_{k=1}^{K} \big[ \epsilon(\tau_k) - \epsilon(\tau_{k-1}) \big]= -s < h(\bar X) - h(\hat Y_K)$,
so the path has points $\hat Y_K, \dots, \hat
Y_0 = \bar X$ with
\begin{equation}
  h(\hat Y_{k-1}) - h(\hat Y_k) > \epsilon(\tau_k) - \epsilon(\tau_{k-1}).
  \label{eq:refseq}
\end{equation}
Let $\hat X_k = M_{\tau_k}^{-1}\hat Y_k$, so that $\hat X_0 = \bar X$
as $M_0 = I$, and
$\tilde h(\hat X_k,\tau_k) = h(\hat Y_k) + \epsilon(\tau_k)$ increases strictly
as $k$ decreases.

Since $\tilde h$ vanishes at $\hat X_K$ it is non-negative at every step, and by
\eqref{eq:refseq} it increases strictly. Hence the unguided update from
$\hat X_k$ to $\hat X_{k-1}$ satisfies \eqref{eq:cbf} with strict inequality,
and Algorithm~\ref{alg:sampler} sets $u_k = 0$ there. The recursion
\eqref{eq:step} adds $g(\tau_k)\sqrt{\Delta\tau_k}\,\zeta_k$ to a term
set by $X_k$ and $u_k$, and $g(\tau_k) > 0$, so exactly one increment
$\zeta^\star_k$ carries $\hat X_k$ to $\hat X_{k-1}$ and
$(\hat X_K, \zeta^\star_{1:K})$ returns $\bar X$.

The barrier $\tilde h$ is continuous and \eqref{eq:refseq} is strict, so
induction on $k$ gives $u_k = 0$ at every step on a neighborhood of
$(\hat X_K, \zeta^\star_{1:K})$. The recursion is the unguided one there, hence
continuous, and a smaller neighborhood is carried into $B(\bar X,\rho')$.
That neighborhood has positive probability under the Gaussian initialization
and increments, so
$\mu(B(\bar X,\rho) \cap \mathcal{C}) \ge \mu(B(\bar X,\rho')) > 0$. If
instead $h(\bar X) = 0$, then $B(\bar X,\rho/2)$ contains a trajectory $X'$
with $h(X') > 0$, because $\mathcal{C}$ is the closure of $\{h > 0\}$, and the
claim follows from $B(X',\rho/2) \subseteq B(\bar X,\rho)$. Finally, we prove the second claim. If \eqref{eq:cbf} holds at every step,
then $h(X_0) \ge 0$ for every realization of the noise, by the extension
of~\cite[Theorem~4.2]{DG-AA-FP:26} given after Lemma~\ref{lem:Mtau}, so that
$\mu(\mathcal{C}) = 1$. Since $h$ is continuous, $\mathcal{C}$ is closed, and
hence $\supp(\mu) \subseteq \mathcal{C}$. The reverse inclusion
$\mathcal{C} \subseteq \supp(\mu)$ is the first claim, since every ball around a
point of $\mathcal{C}$ has positive probability.
\end{proof}

We next ask which agents cooperate, and when. Let
the times at which agent $i$ is admissible only through its coupling constraint be
\begin{equation}
  S_i(X) = \{ t : c_i(x_i(t),u_i(t)) < 0 \le \ccoup{i}(x(t),u(t)) \},
  \label{eq:pattern}
\end{equation}
and we call $S(X) = (S_1(X), \dots, S_N(X))$ the coupling pattern of $X$. The
trajectories in $\mathcal{C}$ with pattern $S$ form the set
$\mathcal{C}_S = \{ X \in \mathcal{C} : S(X) = S \}$, and these sets
partition $\mathcal{C}$. The sampler returns, with nonzero probability, the
coupling pattern of every trajectory that satisfies the constraints strictly.

\begin{corollary}{\emph{\textbf{(Every coupling pattern
        can occur)}}}\label{cor:roles}
  Let the hypotheses of Theorem \ref{thm:complete} hold and let $X$ satisfy
  $h(X) > 0$ and, for every $t$ and $i$, either
  $c_i(x_i(t),u_i(t)) > 0$ or
  $c_i(x_i(t),u_i(t)) < 0 < \ccoup{i}(x(t),u(t))$. Then the sampler
  returns a trajectory in $\mathcal{C}$ with the coupling pattern of $X$ with
  nonzero probability, that is, $\mu(\mathcal{C}_{S(X)}) > 0$.
\end{corollary}

\begin{proof}
The inequalities in the hypothesis are strict and $c_i$, $\ccoup{i}$ and $h$ are
continuous, so there is a $\rho > 0$ such that every trajectory in $B(X,\rho)$
satisfies them and has $h > 0$. By \eqref{eq:pattern} the coupling pattern is
$S(X)$ throughout $B(X,\rho)$, and
Theorem \ref{thm:complete} gives
$\mu(\mathcal{C}_{S(X)}) \ge \mu(B(X,\rho) \cap \mathcal{C}) > 0$.
\end{proof}

\smallskip The constraint functions $c_i$, $\ccoup{i}$ and $\ctask$ determine
$\mathcal{C}$. By Theorem~\ref{thm:complete}, the sampler can return every
trajectory in $\mathcal{C}$, with probabilities determined by the demonstrations. The constraints do not prescribe a coupling pattern, because
$(x(t),u(t))$ lies in $\mathcal{A}_i$ whenever $c_i \ge 0$ or
$\ccoup{i} \ge 0$. Hence $\mathcal{C}$ contains trajectories with different
coupling patterns, and Corollary~\ref{cor:roles} gives each of them a positive
probability of being sampled. Finally, by Proposition~\ref{prop:product}, each
agent contributes its own score to the product prior, while the constraints
enter only through the guidance. Adding or removing an
agent therefore changes only the constraint functions and the set of scores
in the product, and no network needs to be retrained.

The probability that the sampler returns a trajectory in $\mathcal{C}$ depends on
whether \eqref{eq:cbf} can be satisfied at every step, because \eqref{eq:cbf} at every step guarantees that the returned trajectory lies in $\mathcal{C}$.
Algorithm~\ref{alg:sampler} searches for such a step locally, along gradients of
$\tilde h$. Because $h_{\mathrm{comp}}$ in \eqref{eq:hcomp} is a maximum over
the two sets of each union, $h$ can have local maxima at which $h < 0$, and a
search that reaches one may find no step satisfying \eqref{eq:cbf} even when
$\mathcal{C}$ is nonempty. Controllers based on control barrier functions have the same limitation, since their quadratic programs can stall at points other
than the intended ones~\cite{MFR-APA-PT:21}. The maxima that encode unions of sets~\cite{PG-JC-ME:20} add such points.

\section{Numerical results}\label{sec:sim}
We consider Example \ref{ex:ferry} over $50$ timesteps, with $w = 0.16$,
$v^{\max} = 0.045$, $r = 0.05$ and $\delta = 0.06$. The training sets
$\mathcal{D}_1$ and $\mathcal{D}_2$ contain $2000$ single-agent trajectories
each, recorded in an environment of the same dimensions without the river.
Hence the demonstrations describe how each agent moves, and contain neither the
constraints nor any cooperation.\footnote{Each score is a 1-D convolutional
network over time, trained by denoising score matching with
$g(\tau)^2 = 0.1 + 19.9\,\tau$. Algorithm~\ref{alg:sampler} uses $K = 2000$
steps on the grid $\tau_k = (k/K)^2$, $\alpha = 2$ and $j_{\max} = 300$.}
To compute the guidance from the gradient of $h$, we replace the $\min$ and
$\max$ in $h$ by the smooth surrogates
\begin{align*}
  \smin_\kappa(a) &= -\tfrac{1}{\kappa}\log \textstyle\sum_j e^{-\kappa a_j},\\
  \smax_\kappa(a) &= \tfrac{1}{\kappa}\log \textstyle\sum_j e^{\kappa a_j}
                    - \tfrac{1}{\kappa}\log M,
\end{align*}
over the $M$ entries of $a$, with $\kappa = 25$. Both surrogates are bounded above by the exact minimum and maximum, so the smoothed barrier is non-negative only on trajectories in $\mathcal{C}$.

\begin{figure}[t]
\centering
\includegraphics[width=\textwidth]{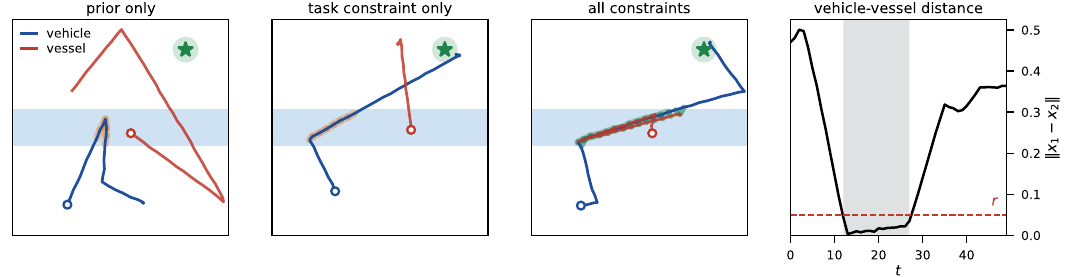}
\caption{River crossing of Example~\ref{ex:ferry}: a trajectory sampled from
the prior alone, one guided by the task constraint $\ctask$ only, and one guided
by all constraints $h$ (Algorithm~\ref{alg:sampler}). The band is the river,
open circles mark initial positions and the disc around $\star$ is the goal. The
vehicle's path in the river is green where the vessel is within $r$ and amber
where it is not. Right: vehicle-vessel distance along the trajectory guided by
$h$, shaded while the vessel carries the vehicle.}
\label{fig:main}
\end{figure}

\begin{figure}[t]
\centering
\includegraphics[width=0.6\textwidth]{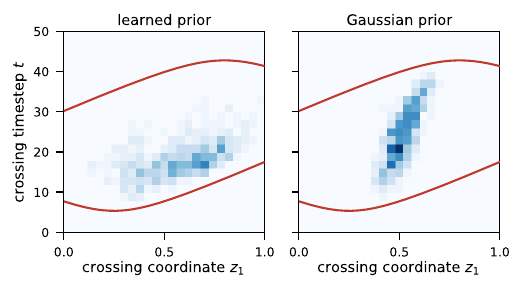}
\caption{$500$ trajectories from the learned prior and $500$ from a
Gaussian prior of matched moments, both guided by $h$.
The shading counts the trajectories that cross at the coordinate $z_1$ and the
timestep $t$ at which the two agents come closest. The red curves bound the
crossings that allow reaching the goal by $T$. The learned
prior spreads its crossings over that region, and the Gaussian
concentrates them near the middle.}
\label{fig:density}
\end{figure}

The sampler generates cooperative trajectories, although neither training set
contains one. No trajectory drawn from the team prior lies in $\mathcal{C}$
(Table~\ref{tab:baselines}). Guided by $\ctask$ alone, all $500$ generated
trajectories reach the goal, but in every one of them the vehicle enters the
water without the vessel. Guided by $h$, all $500$ lie in $\mathcal{C}$.
Fig.~\ref{fig:main} shows one trajectory of each kind. In the trajectory guided by $h$, the vessel meets the vehicle at the river, the two cross together, and the vehicle continues to the goal. This sequence appears in neither training set.

The constraints require the two agents to cross together, and the learned
prior determines where and when they do. To separate the two contributions, we
replace the learned prior by a Gaussian of matched moments and keep the
barrier. For each of the $500$ trajectories guided by $h$, we record the
coordinate $z_1$ and the timestep $t$ at which the two agents come closest
(Fig.~\ref{fig:density}). With the Gaussian prior all trajectories also lie in
$\mathcal{C}$, and the crossings concentrate near the point where the straight
line from $x_1^0$ to $g$ meets the river. With the learned prior, the
trajectories inherit the turns and pauses of the demonstrations, which are
walks through two or three random waypoints at constant speed, and their
crossings spread over most of the region that the horizon allows. The number of timesteps on which the two agents are within $r$ of each other varies in the same way, from $4$ to $27$. The constraints require only two such timesteps, so the longer couplings come from the prior.

The sampler requires no retraining when the team changes. We add a second
vessel that can also carry the vehicle, so that $\mathcal{A}_1$ in
\eqref{eq:union} gains a third set. Both vessels use the same score, and
$\ctask$ fixes the initial position of each, the new vessel starting on the
water at distance $0.2$ from $x_2^0$. All $500$ trajectories generated for this
team lie in $\mathcal{C}$, and in $127$ of them the vehicle passes from one
vessel to the other. With three vehicles and three vessels, all $500$
trajectories lie in $\mathcal{C}$. They show $144$ different coupling
patterns, against $12$ in $100$ solutions of trajectory optimization without
the prior. Removing agents works in the same way, unless the remaining team cannot complete the task. With no vessel, for example, Assumption~\ref{as:coop} fails and no trajectory lies in $\mathcal{C}$.

\begin{figure}[t]
\centering
\includegraphics[width=0.6\textwidth]{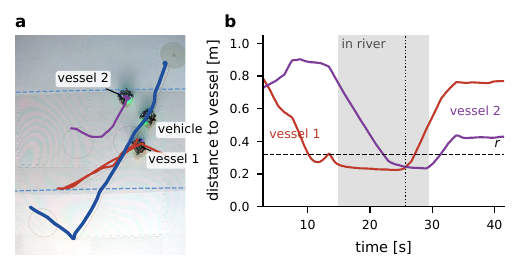}
\caption{
A Robotarium run with two vessels. (a) Camera frame at the
handover, with executed paths of the vehicle (blue), vessel~1 (red) and
vessel~2 (purple). Dashed lines are the riverbanks. (b) Distance from the vehicle to each vessel, with the time in the river
shaded and the time of (a) dotted.
}
\label{fig:robotarium}
\end{figure}

We implement the river crossing on the
Robotarium~\cite{SW-PG-LW-SM-GN-MM-ME:20}, a testbed of unicycle robots in an
arena of $3.2$~m by $2$~m. The unit square of Example~\ref{ex:ferry} is mapped
to a square of side $2$~m in the arena, and each timestep of a trajectory is
executed in $1.2$~s. The robots have a physical size, so we change the
constraint functions and keep the networks and the sampler. The river widens
to $80$~cm and $r$ grows to $32$~cm, so that two robots fit within $r$ of each
other, while $2r < w$ still forces the vessel to carry the vehicle. We also add
two constraints that the simulation does not need. The first requires every
pair of robots to stay at least $20$~cm apart, so that the robots do not
collide. The second requires every robot to stay inside the arena. Finally, a constraint that holds at every timestep can still be violated
between two consecutive timesteps, so we impose every constraint on the whole
segment between them. We call this the Robotarium setting.
Algorithm~\ref{alg:sampler} places every trajectory in $\mathcal{C}$, with one
vessel and with two. Each robot tracks its sampled positions through the
Robotarium's map from single-integrator to unicycle inputs, so the
nonholonomic constraint appears only as tracking error. We execute the first
$15$ trajectories with one vessel and the first $5$ with two. All $20$ runs
satisfy the constraints up to the tracking error, which stays below $5$~cm.
Fig.~\ref{fig:robotarium} shows a run with two vessels, in which one vessel hands the vehicle to the other.

In the second task, an escape room, cooperation also has to be timed. Two
agents start on one side of a wall with a doorway, and both have to reach an
exit on the other side by the end of the horizon. The door is open at time $t$
when, at $t-1$, some agent was on a pad placed before the wall, or in the
doorway. The state of agent $i$ consists of its current and previous
positions, $p_i$ and $p_i^-$, because the door depends on the previous step.
Let $W$ be the wall with the door shut, $W_s \subset W$ its solid part, so that
$W \setminus W_s$ is the doorway, $P$ the pad, and $d_S$ the signed distance
to $S$, positive outside $S$. Then
\begin{align*}
  c_i(x_i,u_i) &= \min\big\{ d_W(p_i),\ v^{\max} - \norm{u_i} \big\},\\
  \ccoup{i}(x,u) &= \min\big\{ d_{W_s}(p_i),\ o(x),\ v^{\max} - \norm{u_i} \big\},\\
  o(x) &= \max_j \, \omega(p_j^-),\\
  \omega(p) &= \max\big\{ {-d_P(p)},\ \min\{d_{W_s}(p), -d_W(p)\} \big\},
\end{align*}
where $\omega(p) \ge 0$ when $p$ is on the pad or in the doorway, so that
$o(x) \ge 0$ when the door is open. The task function $\ctask$ asks each agent
to start within $\delta$ of its initial position and to end within $\delta$ of
the exit. The wall is thicker than one step and the pad is more than one step
from the doorway, so an agent alone cannot pass. Two agents can, one on the pad
while the other enters the doorway and then holds the door. Both agents use the same score network, trained on $4000$ single-agent trajectories recorded in the room without the wall.

\begin{figure}[t]
\centering
\includegraphics[width=0.6\textwidth]{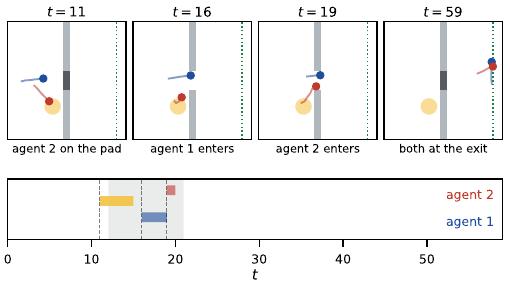}
\caption{A trajectory of the escape room in $\mathcal{C}$, at four timesteps (top) and over the horizon
(bottom). Dots are the agents and faint lines their last five
steps. The doorway is dark when
the door is shut, the disc is the pad, and the dotted line is the exit.
Agent~2 is on the pad from $t = 11$ to $t = 15$, so the door is open from
$t = 12$, agent~1 enters the doorway at $t = 16$ and holds the door, and agent~2
enters at $t = 19$. Bottom: timesteps on the pad (yellow), in
the doorway (agent colour), and with the door open (shaded).}
\label{fig:escape}
\end{figure}

Cooperation that has to be timed is harder to generate. In the escape room,
about $30\%$ of the trajectories generated by Algorithm~\ref{alg:sampler} lie
in $\mathcal{C}$, and in each of them one agent stands on the pad while the
other enters the doorway (Fig.~\ref{fig:escape}). Without the local search,
as in~\cite{DG-AA-FP:26}, no trajectory lies in $\mathcal{C}$. Most failures
leave an agent at the shut door, which is consistent with the local maxima of
$h$ discussed in Section~\ref{sec:main}.

\begin{table}[t]
\caption{Percentage of the generated trajectories that are admissible and
complete the task, averaged over five seeds.}
\label{tab:baselines}
\centering\footnotesize\setlength{\tabcolsep}{5pt}
\begin{tabular}{lccc}
\hline
 & river crossing & escape room & Robotarium\\
\hline
Algorithm~\ref{alg:sampler} & $100$ & $30$ & $100$\\
per-step projection & $100$ & $36$ & $99$\\
penalty guidance & $16$ & $0$ & $14$\\
projection after sampling & $44$ & $2$ & $54$\\
rejection & $0$ & $0$ & $0$\\
\hline
\end{tabular}
\end{table}

In all three settings, Algorithm~\ref{alg:sampler} and per-step projection
place far more trajectories in $\mathcal{C}$ than the other three methods in
Table~\ref{tab:baselines}. All methods use the same prior,
barrier and noise. Per-step
projection~\cite{SZ-JKC-LO-DA-FF:25,JL-JKC-SK-FF:25} moves the sample toward
$\{h \ge 0.02\}$ after every reverse step with a few Newton steps, on $X_\tau$
or on $M_\tau X_\tau$, and the table reports the better of the two. Penalty
guidance~\cite{MJ-YD-JBT-SL:22} adds to the score the gradient of a penalty on
constraint violation, with a weight chosen per task. Projection after sampling
moves each sample of the prior onto $\mathcal{C}$ once sampling ends. Rejection
keeps the samples of the prior that lie in $\mathcal{C}$. These last three
methods place at most about half of the trajectories in $\mathcal{C}$, and
almost none in the escape room. Per-step projection comes with no guarantee
that the returned trajectory lies in $\mathcal{C}$, while
Algorithm~\ref{alg:sampler} has one whenever \eqref{eq:cbf} holds at every
step. In the river crossing and in the Robotarium setting, both methods place
almost every trajectory in $\mathcal{C}$, and projection is faster. In the
escape room, projection places $36\%$ of the trajectories in $\mathcal{C}$,
against $30\%$. It moves the samples about $400$ times more than
Algorithm~\ref{alg:sampler}, as measured by the control energy
of~\cite{DG-AA-FP:26}, and its trajectories end farther from the prior, as
measured by the maximum mean discrepancy~\cite{AG-KMB-MJR-BS-AS:12}. In the
river crossing, projection moves the samples about $50$ times more, and its
trajectories are as close to the prior as those of Algorithm~\ref{alg:sampler}.
On images, the larger correction of per-step projection visibly degrades the
samples~\cite{DG-AA-FP:26}.

\section{Conclusion}\label{sec:conclusion}
In this paper we study how to generate trajectories for a team of agents, without demonstrations of cooperation, when the task requires a capability that no agent has alone. We encode the capability as constraints on the team trajectory, and guide the sampling from diffusion models trained on each agent alone with a barrier function built from the constraints. The resulting sampler can provably generate every team trajectory that satisfies the constraints. Our numerical results show that the sampler generates cooperative trajectories that satisfy the constraints, and that it requires no retraining when agents join or leave the team. On the Robotarium, the generated trajectories satisfy the constraints also when unicycle robots execute them, up to the tracking error. Existing guidance methods, however, can fail when the constraint sets are complex, for example when cooperation also has to be timed. Future work includes the design of guidance methods for complex constraint sets, and learning the coupling constraints from demonstrations of interacting agents.

\section*{Acknowledgments}
This work was supported in part by NSF under award \mbox{CMMI-2622263}.

\bibliographystyle{IEEEtran}
\bibliography{alias,composition-refs2}

\end{document}